\documentclass{ecai/ecai}
\usepackage{times}
\usepackage{latexsym}
\usepackage{amsmath}
\usepackage{amsfonts}
\usepackage{booktabs}
\usepackage{soul}
\usepackage{graphicx}
\usepackage{float}
\usepackage{footmisc}
\usepackage{bm}
\usepackage{balance}
\usepackage{url}

\def\sA{{\mathbb{A}}}
\def\sS{{\mathbb{S}}}
\def\sR{{\mathbb{R}}}
\def\sQ{{\mathbb{Q}}}
\def\sZ{{\mathbb{Z}}}
\def\vw{{\bm{w}}}
\renewcommand{\P}{\mathcal{P}}
\newcommand{\R}{\mathcal{R}}
\usepackage{xcolor}

\title{Distantly Supervised Question Parsing}

\author{Hamid Zafar \institute{\: University of Bonn, Germany, hzafarta@uni-bonn.de} \and Maryam Tavakol \institute{\: TU Dortmund, Germany, maryam.tavakol@tu-dortmund.de} \and Jens Lehmann
 \institute{\: University of Bonn, Germany, jens.lehmann@cs.uni-bonn.de and Fraunhofer IAIS,
Germany, jens.lehmann@iais.fraunhofer.de}}

\begin{document}
\maketitle

\begin{abstract}
The emergence of structured databases for Question Answering (QA) systems has led to developing methods, in which the problem of learning the correct answer efficiently is based on a linking task between the constituents of the question and the corresponding entries in the database. 
As a result, parsing the questions in order to determine their main elements, which are required for answer retrieval, becomes crucial. 
However, most datasets for QA systems lack gold annotations for parsing, i.e., labels are only available in the form of (question, formal-query, answer). 
In this paper, we propose a distantly supervised learning framework based on reinforcement learning to learn the mentions of \emph{entities} and \emph{relations} in questions. 
We leverage the provided formal queries to characterize delayed rewards for optimizing a policy gradient objective for the parsing model. 
An empirical evaluation of our approach shows a significant improvement in the performance of entity and relation linking compared to the state of the art. We also demonstrate that a more accurate parsing component enhances the overall performance of QA systems.

\end{abstract}

\section{Introduction}
Over the last few years, the emergence of large-scale Knowledge Graphs (KGs), such as DBpedia~\cite{swj_dbpedia} and Freebase~\cite{bollacker2008freebase}, provided well-structured information for question answering (QA) systems. 
This allows for developing QA models that are able to query or reason over such knowledge graphs to support complex questions (such as multi hop questions, aggregation and ordinal questions~\cite{abdelkawi2019complex}) going beyond the types of questions which can usually be successfully answered using unstructured text corpora. 
The mainstream approaches in the field of Knowledge Graph based Question Answering (KGQA) systems are semantic parsing based methods and end-to-end neural network based approaches. Semantic parsing approaches aim to transform a natural language question to an admissible query against the knowledge graph, whereas end-to-end systems directly return results without using a formal query language.
Although recent advances in QA research have been beneficial for developing end-to-end systems~\cite{chakraborty2019introduction}, these approaches have several shortcomings. First, the obtained models from such end-to-end systems are not interpretable, which also means conducting error analysis and improving the models based on intermediate outputs is very difficult. 
Second, other approaches cannot re-use the internal - sometimes very sophisticated - components of these systems. This also restricts optimization using intelligent pipeline composition~\cite{singh2018reinvent}. 
Third, the required amount of training data can be very high.

\begin{figure*}[t]
\centering
\includegraphics[width=0.9\textwidth]{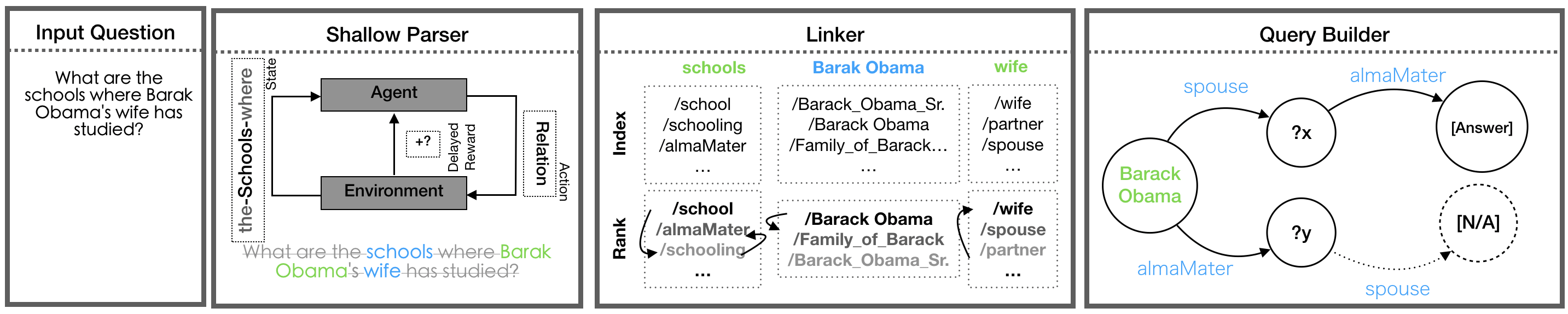}
\caption{The overall pipeline of question answering system over knowledge graph.}
\label{fig:pipeline}
\end{figure*}

Therefore, we aim to study modular question answering systems over knowledge graphs (KGQA) and focus on the (shallow) parsing task, which is usually the essential first component of a typical KGQA pipeline (cmp. Figure~\ref{fig:pipeline}). 
Although shallow parsing is a key step in the process, and almost all recent approaches for complex QA tasks over KGs use shallow parsing, it has received little attention in previous works. The problem is addressed either by a simple technique for pattern/template matching~\cite{dubey2016asknow,sakor2019} or an existing shallow parser from the Natural Language Processing (NLP) community~\cite{dubey2018earl}. 
However, the well-known shallow parsing methods from NLP such as~\cite{akbik2018coling,Collobert2011} are usually not suitable for QA systems. 
A core reason for this is the difference in the syntactical structure of questions compared to normal sentences. Most text corpora contain a low percentage of questions and, naturally, the models trained on those are not properly tuned for questions.
Additionally, these methods are designed to provide a fine granularity level of parsing in the sentences, while we require a more shallow decomposition. 
This motivates designing a proper shallow parser specific for QA systems.
\looseness=-1

\begin{figure}[t]
\centering
\includegraphics[width=0.4\textwidth]{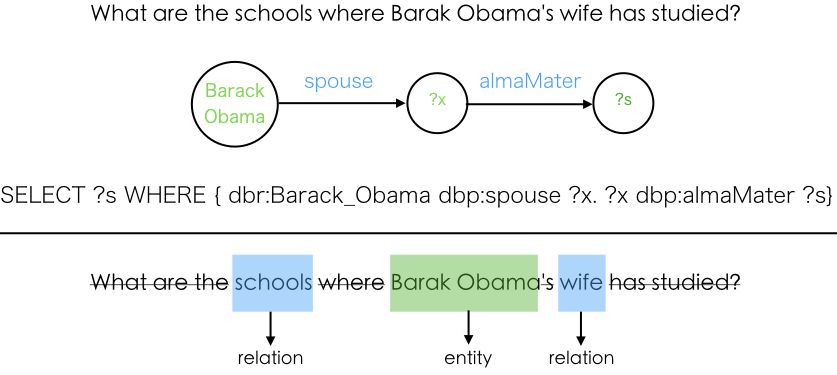}
\caption{An example of shallow parsing task for a given question (with a small typo) and its corresponding formal query in a knowledge graph.}
\label{fig:example}
\end{figure}

Figure~\ref{fig:example} (top) depicts an exemplary (question, formal-query) pair followed by the desired annotations (bottom) that provide the most relevant information required for answer retrieval. 
The output of parsing is further used in the next step of the overall procedure of question answering over KGs which is displayed in Figure~\ref{fig:pipeline}. 
The figure shows that first, a natural question is passed to the shallow parser for detecting the entities and relations of the question. 
Once the words are labeled, the linker finds the corresponding entities/relations from the underlying knowledge graph.
Finally, the top candidates are used to generate a formal query given the entity and relation mentions to retrieve the corresponding answer from the knowledge graph. 
Therefore, the goal of shallow parsing is to annotate the words/phrases of the input questions with appropriate labels, which later will be employed in the entity/relation linking step. 

The above task resembles a standard classification problem in which every entity/relation mention of the input question is required to be labeled according to its role in fetching the corresponding target entity/relation. 
However, to the best of our knowledge, almost all existing question-answer datasets do not provide the target annotations for parsing 
(there is only one such dataset according to \cite{chakraborty2019introduction}). 
Consequentially, supervised learning approaches (including well-known NLP parsers) fall short in addressing the shallow parsing problems in QA systems over KGs. 
Nonetheless, since the target entities/relations are the building blocks of the formal query, we can exert the performance of the linker module as a distant measure for the evaluation of the parser. 
For instance, in the given example of Figure~\ref{fig:example}, if the parser only labels ``\textit{Barak}" as an entity and not ``\textit{Barak Obama}", the linker may not be able to accurately retrieve the corresponding target entity, thus the lower performance of the linker can be interpreted as a distant measure to evaluate the performance of the parser.\looseness=-1

Hence, we propose a novel sequential approach to model the task of distant supervision in a Reinforcement Learning (RL) framework where the long-range loss determines a delayed reward signal. Although reinforcement learning has demonstrated successful performance in many complex tasks~\cite{mnih2013playing,tavakol2014factored}, it has been insufficiently explored in NLP domain. Some examples are using RL in text summarization~\cite{lee-lee-2017-automatic}, structure induction~\cite{yogatama2017learning}, and dialog systems~\cite{singh2000reinforcement,li-etal-2016-deep}. 
The cognitive process of human to analyze a question (or a sentence in general) is sequential in nature, and in our case of English language, from left to right.  More precisely, while a question is read from left to right, after spotting the first entity/relation mention, identifying the subsequent entities/relations becomes easier by relating the next words to the previously detected parts. 
Reinforcement learning approaches are able to perfectly replicate this natural behavior. 
We thus model the task of shallow parsing in QA systems over KGs in a Markov Decision Process (MDP) framework in which the labels can be only obtained at the end of the sequence and are equivalent to the delayed reward in RL scenarios. 
Therefore, the MDP setting copes with the distantly supervised problem, by utilizing the distant metrics as the reward of a complete question. We further employ a policy gradient optimization technique to learn an optimal policy for labeling the constitutes of the questions in this framework.

Empirically, our model exhibits excellent performance compared to the state-of-the-art methods in the tasks of entity/relation recognition on three popular benchmark datasets.
Furthermore, we demonstrate that our approach can be easily integrated into existing entity/relation linking tools and enhances their accuracy, which leads to improving the overall performance of the QA systems. 
The remainder of the paper is structured as follows. 
The next section presents the related work on entity/relation detection and linking approaches, as well as related techniques used in QA systems.  
We introduce the details of our proposed approach in Section~\ref{sec:approach} followed by the empirical study in Section~\ref{sec:results}, and Section~\ref{sec:conc} concludes our contributions.

\section{Related Work} \label{sec:rw}
Traditionally, entity linking and relation linking problems are addressed independently in different applications. There are many works that solely focus on entity recognition and linking~\cite{hoffart2011robust,mendes2011dbpedia,ferragina2012fast,moro2014entity,speck2014ensemble,yang2016s}. These approaches exert a wide range of techniques and methods ranging from heuristics and rule-based methods to supervised learning, and most recently deep neural networks. Shen et al.~\cite{shen2015entity} provide a survey on various approaches and tools for entity disambiguation and linking and their applications in different fields such as question answering. They conclude that the process of entity linking commonly consists of two main steps: generation of candidate entities, in which irrelevant items are filtered out, and a ranking method to identify the best candidates.

On the other hand, several other approaches exist that concentrate only on the task of relation detection as well as linking~\cite{mulang2017matching,gerber2011bootstrapping,nakashole2012patty}. In contrast to relation extraction that have been an active field of research in natural language processing domain~\cite{zelenko2003kernel,mintz2009distant,fundel2006relex}, relation linking has received less attention. However, relation linking has an important role in various applications e.g., question answering, knowledge graph completion, etc. Legacy methods such as PATTY~\cite{nakashole2012patty} and BOA~\cite{gerber2011bootstrapping} provide natural language patterns and taxonomy to create different representations of the relations in the knowledge graph. 
These approaches are additionally employed in question answering systems~\cite{lehmann2012deqa,unger2012template,Zou2014,dubey2016asknow}.
\looseness=-1

Nevertheless, recent works attempt to exploit the correlation between entities and relations by jointly linking them in the questions in order to improve the performance of the state of the art methods~\cite{dubey2018earl,sakor2019}. Dubey et al.~\cite{dubey2018earl} propose an approach for entity and relation linking, called EARL, which employs SENNA~\cite{Collobert2011} to extract the entity and relation spans of the questions. 
They map the task of entity/relation linking to generalized travelling salesman problem, which results in exploiting the connection between entities and relations in the questions. 
Although SENNA attained the state of the art results in multiple tasks, its performance in QA is inferior as it was trained on long text documents from Wikipedia. Whereas the questions in QA datasets are usually short; for instance the average length of questions  in LC-QuAD~\cite{trivedi2017coprus} dataset is 12.3. Falcon~\cite{sakor2019} is based on a set of manually crafted rules and benefits
from general structures of English language to detect the entity/relation mentions. They further merge multiple existing knowledge graphs to increase the lexical diversity of the KG. 
However, their approach does not generalize well on other datasets.

The usage of RL in various question answering as well as dialog systems has been exploited~\cite{zhong2017seq2sql,huang2018natural,shah2018building,andreas2016learning}. For instance, Zhong~\cite{zhong2017seq2sql} proposed an RL-based model to map questions to SQL query using pointer networks. However, as they need large amount of training data, they used crowd sourcing
to generate natural language question for the automatically created queries.
A similar work to our research is presented by Takanobu et al.~\cite{takanobu2019hierarchical} in the NLP domain, where the main objective is to identify the relations. They leverage a two level RL-based framework to first detect a relation in the top level, which triggers the lower level to find the corresponding entities that are semantically associated to the selected relation. Their proposed method is evaluated on two very small datasets (total of 29 and 11 relation). 
Nevertheless, the approach is presented for scenarios in which the gold-annotations are available which makes it unsuitable to compare to our approach and the reward is computed from the prediction error.\looseness=-1





\section{Distant Supervision Model}\label{sec:approach}
In this section, we present our approach for shallow parsing of questions in natural language for QA systems that operate over knowledge graphs. 
The task is phrased as a reinforcement learning method to model the problem of distant supervision and is thus formulated in a Markov Decision Process (MDP) framework. 
\subsection{Preliminaries}
We are given a set of $N$ questions $\sQ=\{q_i,\ldots,q_N\}$ and a set of their corresponding formal queries $\sZ=\{z_1,\ldots,z_N\}$. 
Each question $q_i\in\sQ$ consists of a sequence of $n_i$ words, $q_i=[\vw_1^i,\vw_2^i,\ldots,\vw^i_{n_i}]$, which are determined with an arbitrary vectorized representation of size $d$, $\vw_j^i\in\sR^d$. 
On the other hand, the formal queries are the source of quantifying the so-called distant labels in our setting. 
A query $z_i\in\sZ$, corresponding to $q_i$, is formed from a set of $m_i\leq n_i$ linked items, $z_i=\{l^i_1,l^i_2,\ldots,l^i_{m_i}\}$, where a linked item $l_j^i$ is defined as a triplet of (\textit{title}, \textit{URI}, \textit{label}) that links a part of the question (one or more words), i.e., \textit{title}, to a \textit{URI} entry in the knowledge graph with a \textit{label} of either \textit{relation} or \textit{entity}. 
For instance, given the example question of Figure \ref{fig:example}, the linked items are as follows 
{\small
\{(\textit{``almaMater"}, \texttt{dbp:almaMater},  \textbf{relation}),
(``\textit{Barack Obama}", \texttt{dbr:Barack\_Obama},
\textbf{entity}),
(``\textit{spouse}", \texttt{dbp:spouse}, \textbf{relation})\}}.\looseness=-1

We aim to design a parsing method, which receives a question $q_i\in \sQ$ and identifies its entities and relations by classifying every word into entity, relation, or none. 
Therefore, every word $\vw_j^i$ in the question $q_i$ is assigned to a label $y_j^i \in \{2,1,0\}$ which stand for entity, relation, and none, respectively.
Consequentially, the output vector of the parser for the question ``\textit{What are the schools where Barak Obama's wife has studied}" is $y^i=[0,0,0,1,0,2,2,0,1,0,0]$.

\subsection{The MDP Framework}
We model the task of question parsing as a sequential approach based on Reinforcement Learning (RL). 
The cycle of learning in an RL  problem consists of an agent perceiving the state of the environment, performing an action,  accordingly, and the environment provides a feedback to evaluate its action. The (delayed) feedback thus operates as a reinforcing signal for optimizing the internal model of the RL framework. 
An RL problem is mathematically described via Markov Decision Processes (MDPs) \cite{sutton1998introduction}. 
An MDP is represented via a five-tuple $\langle \sS,\sA,\P,\R, \gamma \rangle$, where $\sS$ is the state space, $\sA$ is the set of actions, $\P : \sS \times \sA \times \sS \rightarrow [0,1]$ is the transition probability function where $\P(s_t,a_t,s_{t+1})$ indicates the probability of going to $s_{t+1}$ after taking action $a_t$ in state $s_t$ at time $t$, in which $s_t,s_{t+1} \in \sS$ and $a_t\in\sA$, $\R : \sS \times \sA \rightarrow \sR$ is the reward of a state-action pair, and $0< \gamma\leq 1$ is the discount factor. 
The goal of an MDP is to learn a policy $\pi:\sS \times \sA \rightarrow [0,1]$ which maximizes the expected obtained reward. A stochastic policy $\pi$ gives a probability distribution over the possible actions that the agent can take in the current state. 

In our setting, we assume that the input questions are equivalent to the episodes in the reinforcement learning framework. 
The agent traverses the question from left to the right and decides to choose a label for every word in the sequence based on the information encoded in the current state. At the end of the episode, the obtained labels are integrated and are used to compute a distant loss value which forms a delayed reward signal. In the remainder of this section, we characterize our shallow parsing problem in an MDP framework. 
Moreover, we discard the superscript $i$ for brevity in the notation and denote the time step within the episodes of the MDP by index $t$.\looseness=-1

\textbf{States. }In our setting, the state space is defined as a subsequence of the question at each time step $t$ as well as the last chosen action. We introduce a parameter $h$ to control the size of state by considering a window of $2h+1$ words over the input question. A state $s_t$ thus encodes the current word $\vw_t$, $h$ previous words, $h$ next words, and the previous selected action, $s_t=[\vw_{t-h},\dots,\vw_t,\dots,\vw_{t+h},a_{t-1}]$. Consequently, taking action $a_t$ at this time step will lead to the next state $s_{t+1}=[\vw_{t-h+1},\dots,\vw_{t+1},\dots,\vw_{t+h+1},a_{t}]$.

\textbf{Actions.}
We aim to find the mentions of relations and entities of the questions by classifying their words into a possible set of three labels. 
We hence specify a discrete action space of $\sA=\{0,1,2\}$, where $a_t \in \sA$ determines whether the current word is an entity or a relation, or it is out of our interest. At the end of the episode (question), the selected actions, $A_q= [ a_1,\ldots, a_n ]$, form a sequence of mentioned labels for the corresponding words. 
Note that an action is equivalent to a predicted label, i.e., $a_t=\hat{y}_t$.

\textbf{Transition function.}
As choosing an action would only lead to one possible next state, the transition function is deterministic in our problem. 
That means, $\P(s_t,a_t,s_{t+1})=1$ for $s_{t+1}=[\vw_{t-h+1},\dots,\vw_{t+1},\dots,\vw_{t+h+1},a_{t}]$ and is zero otherwise.


\textbf{Reward function.}
Since the true labels of the words are not given in every state, we are able to evaluate the selection policy only at the end of the question. Therefore, no immediate reward is available at each time step. We thus delay the policy evaluation and utilize the assessment of the linker to compute a distant score for our prediction which serves as a delayed reward for updating the policy. We describe our method for computing the distant feedback below.\looseness=-1


\subsection{The Distant Labels and Reward}
Consider the question provided in Figure~\ref{fig:example} again. 
The example depicts certain complexities in the parsing task that we should take into account. 
An entity or a relation could refer to more than one word. 
For instance, the entity \texttt{dbr:Barack\_Obama} is specified by separately labeling two words \textit{Barak} and \textit{Obama} as entity. 
Furthermore, as the example illustrates, there is not always  a one-to-one mapping between a linked item and a word (or set of words) in the question, e.g., \textit{wife} vs. \texttt{dbp:spouse} and \textit{schools} vs. \texttt{dbp:almaMater}. 
In addition, recall that the true labels, i.e., entity/relation mentions in the questions, are not available in our problem.\looseness=-1

We propose to learn a policy for labeling every word $\vw_t$ of a question $q$ from a distant reward value which is computed from the quality of linked items $l\in z$. As a result, our RL framework becomes a distantly supervised approach for the underlying parsing task.
To do so, we find the phrases (sequence of words) from the question $q$ which provide the best indication for each target linked item $l_j$. 
Let $a_t^\pi$ be the action (aka label) chosen by policy $\pi$ for word $\vw_t$ and assume that $a_t^\pi$ is either entity or relation, i.e., $a_t^\pi \neq 0$, we group the words within the same label to construct the set of entity and relation mentions for the question at-hand. 
For instance, if $(a_{t-1}^\pi\neq a_t^\pi) \land (a_t^\pi = a_{t+1}^\pi = \dots  = a^\pi_{t+b} ) \land (a^\pi_{t+b} \neq a^\pi_{t+b+1})$, we concatenate these sequence of $b$ words into one phrase, $\omega_k=[\vw_t , \dots ,\vw_{t+b}]$ with label $\lambda_k=a_t^\pi$. Hence, at the end of the episode (question), a set of $g$ entity and relation mentions $\Omega=\{{\omega_1,\dots,\omega_{g}}\}$ are obtained along with their predicted labels $\Lambda=\{{\lambda_1,\dots,\lambda_{g}}\}$, where  $g \leq n$, and optimally $g=m$.
Using a well-qualified similarity function $\phi(.,.)\mapsto{\sR}$, we aim to compute a score for prediction of each $(\omega_k,\lambda_k)$ pair from the actual policy by finding a linked item from $z$, which is of the same label as $\lambda_k$ and its title is the most similar to $\omega_k$
\begin{equation}\label{eq:score}
    \text{score}(\omega_k, \lambda_k)=\max_{l\in z ,  \text{ label}(l)=\lambda_k} {\phi(\omega_k, \text{title}(l))}.
\end{equation}
As a result, $\text{score}(\omega_k, \lambda_k)$ computes how relevant the predicted label for the phrase $\omega_k$ is to the label of the most similar target linked item, and will later be used to define the delayed reward of an episode/question. 
Continuing with the question in the earlier example, if the model labels the word \textit{``where"} as a relation, it would get a very low score as it is not similar to neither of both target relations. In contrast, if the word \textit{``Obama"} is marked as an entity, it will be matched to \texttt{dbr:Barack\_Obama} with a fairly high score, and when the model correctly identifies both \textit{``Barak"} and \textit{``Obama"} as an entity phrase, we get an almost exact match. 
Note that the choice of similarity function is very essential as it should be able to detect the words with the same meaning, such as \textit{``spouse"} and \textit{``wife"}, as well as typographical errors, for instance the letter \textit{``c"} is missing in \textit{``Barak"}.\looseness=-1

Now we utilize the obtained scores to define a distant feedback for the whole episode which provides a delayed reward.
Given the scores computed from Equation~(\ref{eq:score}), the total reward is defined as the average scores of all identified phrases in the question  
\begin{equation}\label{eq:rew}
 r = \frac{1}{g}\sum_{k=1}^g \text{score}(\omega_k,\lambda_k),
\end{equation}
where $r$ also represents the so-called distant measure for the whole question.
The obtained reward is further discounted by the factor $\gamma$ for the previous states and is used to specify feedback for the overall episode, particularly, the words that are not identified as any entity/relation mention and labeled as zero. 

\subsection{Optimization}
Once the shallow parsing task is successfully phrased in an MDP framework, we aim to learn a deterministic policy $\pi$ which provides the best action (annotation) given the actual state (word). 
Note that the learned policy in our RL framework is a stochastic policy, and we can simply turn that into a deterministic action selection using $\pi(s)=\arg\max_a \pi(a|s)$. 
Subsequently, we take a policy gradient method~\cite{sutton2000policy} to directly learn the optimal policy without learning the intermediate value functions. 
The goal of policy gradient algorithm is to learn a policy $\pi$  with parameters $\theta$, i.e., $\pi_\theta$, where following that policy maximizes the ``expected" obtained reward. 
Recall that in our setting, the episodes are formed from the available questions in the training data. However, following various policies creates different episodes from a single question which leads to a total of $M$ episodes. We thus assume that an episode $\tau_i$ is formed when following policy $\pi_\theta$ with a probability of 
\begin{equation*}
    P(\tau_i;\pi_\theta)=P(s_1^i)\prod_t \pi_\theta(a_t^i|s_t^i)\P(s_t^i,a_t^i,s_{t+1}^i),
\end{equation*}
and since the transition probability is deterministic, we have
\begin{equation*}
    P(\tau_i;\pi_\theta)=P(s_1^i)\prod_t \pi_\theta(a_t^i|s_t^i).
\end{equation*}
Let $r^i$ be obtained from Equation~(\ref{eq:rew}) and  $R(\tau_i)=\sum_t \gamma^{t-1} r^i_t$ be the total discounted reward acquired for episode $\tau_i$ while following policy $\pi_\theta$, the objective function is defined as 
\begin{equation}\label{eq:obj}
    J(\theta)=\sum_{i=1}^M P(\tau_i;\pi_\theta) R(\tau_i),
\end{equation}
which aims to maximize the expected reward over all possible episodes, weighted by their probabilities under policy $\pi_\theta$. 
Hence, we estimate the gradients in the direction of higher discounted reward to update the parameters $\theta$ of the policy $\pi$ via gradient ascent
\[\theta \leftarrow \theta + \alpha \nabla_\theta J(\theta),\]
where $\alpha$ is the learning rate, and by taking the log of likelihood, the gradient becomes
\begin{equation*}
    \nabla_\theta J(\theta)=\sum_{i=1}^M \nabla_\theta \log \pi_\theta(a_t^i|s_t^i).
\end{equation*}

Given the objective function in Equation~(\ref{eq:obj}), we employ a deep learning method as a general function approximation technique to learn the parameters $\theta$. 
We thus design a fully connected neural network with three layers as the policy network. In the first layer, the words are vectorized using a word vectorization technique e.g., word embedding~\cite{mikolov2013distributed,pennington2014glove}. The state is then created using the current, the $h$ previous and $h$ next vectorized words along with the last chosen action, which is a numerical value. 
The second layer can be either a linear transformation with Relu as an activation function, or an LSTM or a Bi-LSTM. We compare different architectures in Section~\ref{sec:results}. 
The output layer uses Softmax activation to provide the final action distributions, from which the agent would sample the next action
$a_{t+1} \sim \pi_{\theta}(s_t,a_t)$.


\section{Empirical Study} \label{sec:results}
In this section, we evaluate the performance of our proposed approach for shallow parsing in QA systems over KGs. 
We conduct our experiments on three popular benchmarking datasets: (i) LC-QuAD~\cite{trivedi2017coprus}, (ii) QALD-6~\cite{unger20166th}, and (iii) QALD-7~\cite{usbeck20177th}. 
The first dataset consists of 5,000 manually crafted question-query pairs that covers a wide range of vocabulary, semantic complexity, spelling and grammar mistakes, and types of question. The other two datasets are from the Question Answering over Linked Data (QALD~\footnote{http://qald.aksw.org/}) challenge series. QALD-6 contains 350 training questions as well as 100 questions in the test set, whereas QALD-7 has 215 training questions and 50 questions in the test set.

Recall that in our distantly supervised setting, the target entity/relation mentions are not available. 
Therefore, we evaluate our parser in combination with a linker to the KGs as a proxy for assessing its performance compared to the baseline linking methods. In the remainder, we denote the combination of our parser and the linking component by \textit{MDP-Parser}.
The performance of different methods is evaluated via \textit{accuracy} which indicates whether the top candidate obtained from our method correctly matches the items identified in the target query. Furthermore, we employ \textit{Mean Reciprocal Rank (MRR)} in cases where a list of candidates with length $k$ are presented $ \text{MRR} = \frac{1}{k}\sum_{i=1}^{k} \frac{1}{\text{rank}_i},$
where $\text{rank}_i$ specifies the position of the target item in the list of candidates.
In addition, we employ a Bayesian optimization approach using SigOpt~\cite{pmlr-v84-martinez-cantin18a} for hyper-parameter optimization on the validation set.\looseness=-1
\paragraph{Reproducibility.} The source code and all intermediate files of our method is available on \url{https://github.com/AskNowQA/DeepShallowParsingQA} to facilitate reproducibility.

\subsection{Linking Component}\label{sec:link}
We design a simple linking component which is able to retrieve the entities and relations from the knowledge graph given the output of the shallow parsing. 
The linking component consists of two steps: retrieving the candidate items and ranking them. In the first step, we employ Apache Lucene~\footnote{https://lucene.apache.org} to create a full-text index of the character-level tri-grams of the entities/relations from the underlying KG. 
These obtained indices provide a list of candidate URIs given the mentions that parser predicted to be either an entity or a relation (see Figure~\ref{fig:pipeline}). 
Note that we distinguish \textit{``relation"} linking from \textit{``entity"} linking in the evaluation as they have distinct characteristics. 
Furthermore, in order to increase the accuracy of the relation linking, we extend the output of the relation index with the relations that are in a one-hop distance of the top candidate entity for each entity mentions. 
In the second step, the candidates are ranked according to a pair-wise ranking function which utilizes an appropriate similarity method.\looseness=-1

We consider three different techniques for computing the similarity in the ranking function that capture either the semantic similarity, surface resemblances, or both. 
In order to capture the string similarity, we examine the top-10 best performing techniques for full name matching reported by~\cite{christen2006comparison}, and we choose levenshtein distance (LEV) to capture the string similarity of a given mention and the corresponding candidates, which performs the best in our setting. 
Second, to support semantic similarity, we utilize GloVe vectors~\cite{pennington2014glove} to initialize the word embedding (EMB) vectors, where we decide for cosine similarity over euclidean distance and word mover's distance~\cite{kusner2015word} due to its efficiency in terms of performance and time. 

Finally, we combine both of them via a linear function (LEV+EMB). 
In the experiments, we use LEV for the entity linking and LEV+EMB for the relation linking, which intuitively means that the string similarity is sufficient for matching the entities, while for the relations semantic similarity is also required.



\subsection{Baseline Methods}
We evaluate the performance of our approach in comparison to four categories of baseline methods. 
First, we employ two well-known NLP parsers, namely SENNA~\cite{Collobert2011} and Flair~\cite{akbik2018coling}, combined with our linking component (see Section~\ref{sec:link}) to measure the gain in the performance of our parser, which is designed for QA systems, compared to an existing NLP shallow parser. 
Note that, to the best of our knowledge, no parser which is specifically designed for QA systems is available to be used as other baselines. 
Second, we compare our approach to an LSTM as well as a Bi-LSTM model which converts the obtained scores into weak labels to address the parsing problem as a classification task. This helps to highlight the importance of
modeling the distant supervision problem in an RL paradigm in contrast to using local scores.


The third group of baselines includes the state of the art entity/relation linking methods that already contain an internal parser. 
We take EARL~\cite{dubey2016asknow} and Falcon~\cite{sakor2019} that report results on both entity as well as relation linking tasks. 
In addition, Babelfy~\cite{moro2014entity}, 
FOX~\cite{speck2014ensemble}, TagMe~\cite{ferragina2012fast} and DBpedia Spotlight~\cite{mendes2011dbpedia} serve as baselines for entity linking, and SIBKB~\cite{singh2017capturing} and ReMatch~\cite{mulang2017matching} are used for the task of relation linking. 
Finally, in order to show the effectiveness of our approach in combination with existing entity/relation tools, we substitute the shallow parsing component in EARL with our parser and address this baseline as \textit{EARL+MDP-Parser}. In general, our parser could be re-used in different methods and combined with various linkers to form a powerful pipeline.

\begin{figure}[t]
\centering
\includegraphics[width=0.30\textwidth]{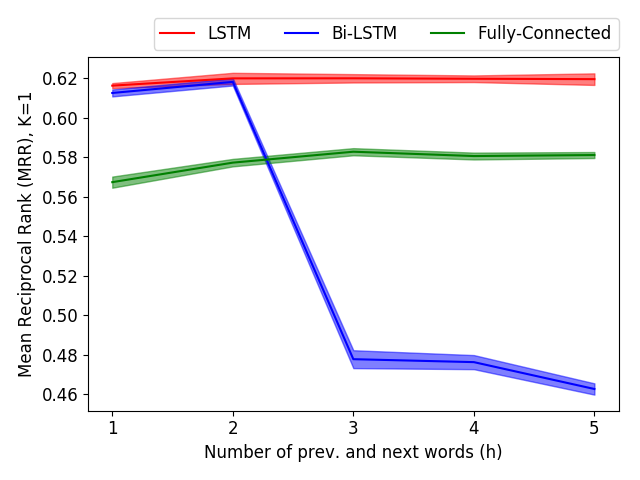}
\caption{Performance in terms of MRR for various network architectures w.r.t. state size on LC-QuAD dataset.}
\label{fig:b_param}
\end{figure}

\subsection{Performance Results}
\subsubsection{Model Selection}
In the first experiment, we analyze the impact of two of the main design factors of our model: the state size which is controlled by parameter $h$ encoding the number of previous and next words included in the state, and the architecture of the deep model.

Figure~\ref{fig:b_param} reveals the effect of different choices of models on the performance of entity and relation  linking task in terms of MRR where $k=1$. 
The experiments are repeated for several runs and report the average results with standard errors. 
The figure illustrates that the LSTM model gives better results than the other models regardless of the value of  parameter $h$. 
Additionally, the state size of 3, i.e.~$h=2$, achieves the best performance in different deep models, however comparing to $h=1$, the improvement is marginal, but the computational costs are much higher. 
Hence, in the remainder of this section, we set $h=1$ and use the deep architecture that contains an LSTM layer for our empirical study.

\begin{table}
\centering
\caption{Accuracies for \textit{entity} linking task.}
\label{tbl:compare_entity}
{\small
\begin{tabular}{@{}llll@{}}
\toprule
Approach & LC-QuAD & QALD-6 & QALD-7\\ \midrule
SENNA & 0.27 & 0.22 & 0.22 \\ 
Flair & 0.59 & 0.56 & 0.43 \\\midrule
LSTM  & 0.47 & 0.43 & 0.46 \\ 
Bi-LSTM  & 0.39 & 0.25 & 0.15 \\ \midrule
Babelfy & 0.17 & 0.29 & 0.29 \\ 
FOX   & 0.40 & 0.49 & 0.41  \\
TagMe & 0.30 & 0.34 & 0.41 \\
EARL & 0.60  & 0.50 & 0.56 \\
DBpedia Spotlight & 0.54  & 0.68 & 0.62 \\
Falcon & 0.74  & \textbf{0.70} & 0.38 \\\bottomrule
MDP-Parser & \textbf{0.76} & \textbf{0.70} &  \textbf{0.69}   \\ 
EARL+MDP-Parser & 0.71 & 0.61 &  0.62 \\
\end{tabular}}
\end{table}

\subsubsection{Evaluation of Entity/Relation Linking}
We first evaluate the performance of various approaches for the task of entity linking. 
Table~\ref{tbl:compare_entity} presents the obtained results on the test-set of LC-QuAD, QALD-6 and QALD-7 datasets. 
The table demonstrates that our MDP-based method outperforms different baseline systems in LC-QuAD and QALD-7 datasets.  
Although Falcon has a good performance on two datasets, it has significantly lower performance on QALD-7 data\footnote{The results of Falcon are computed via: https://labs.tib.eu/falcon}. 
The reason for this lies in the fact that Falcon is a rule-based system which requires manual rule-extraction, and the authors reportedly used other data to extract the rules. 
The low accuracy of 38\% leads to the conclusion that the extracted rules are overfitted and do not generalize well to other datasets. 
Additionally, looking closer to the result of EARL+MDP-Parser illustrates that our method significantly increases the performance of EARL by more than 10\% on LC-QuAD as well as QALD-6, and by a factor of 6\% on QALD-7 dataset.

\begin{table}
\centering
\caption{Accuracies for \textit{relation} linking task.}
\label{tbl:compare_relation}
{\small
\begin{tabular}{@{}llll@{}}
\toprule
Approach & LC-QuAD & QALD-6 & QALD-7\\\midrule
SENNA  & 0.10 & 0.02 & 0.06 \\ \midrule
LSTM  & 0.28 & 0.20 & 0.15 \\
Bi-LSTM  & 0.25 & 0.06 & 0.03 \\\midrule
SIBKB*  & 0.14 & N/A& N/A \\ 
ReMatch* & 0.16  & N/A& N/A \\
EARL &  0.26 & 0.18 & 0.15  \\ 
Falcon &  0.38 & 0.30 & 0.13  \\ \bottomrule
MDP-Parser & \textbf{0.45} & \textbf{0.34}&  \textbf{0.25}  \\ 
EARL+MDP-Parser & 0.34 & 0.22 & 0.24 \\ 
\end{tabular}}
\\
* Results are taken from~\cite{sakor2019}
\end{table}

We further evaluate the performance of our method compared to the baselines for the task of relation linking. 
Table~\ref{tbl:compare_relation} summarizes the results in terms of accuracy. Despite the simplicity of the linking component in our approach, MDP-Parser achieves better results compared to the baselines, and again, improves the performance of EARL by almost 5\% on all the datasets. 
Note that the overall low performance of relation linking methods in comparison to the entity linking is due to the intrinsic complexity in various dimensions, such as difficulty of capturing the relation mentions, semantic closeness, similar candidates from the underlying KG, etc.


\begin{figure}[t]
\centering
(a) Entity linking
\\
\includegraphics[width=0.30\textwidth]{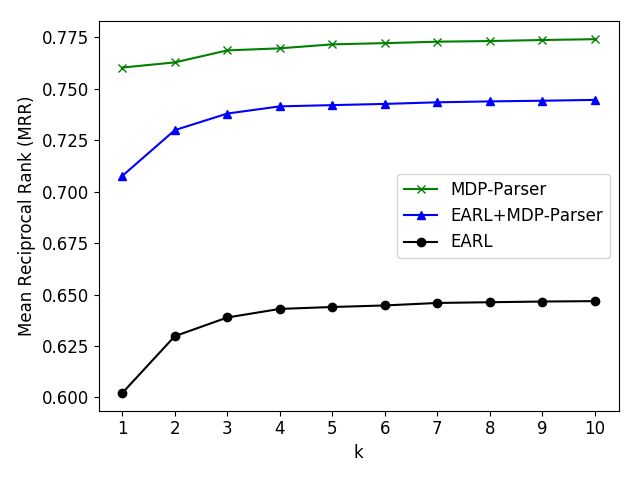}\\
(b) Relation linking\\
\includegraphics[width=0.30\textwidth]{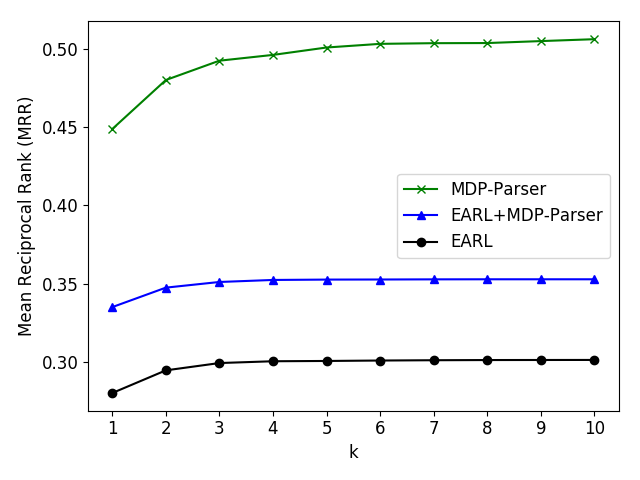}
\caption{Performance in terms of MRR w.r.t. the value of $k$.}
\label{fig:mrr_k}
\end{figure}

Moreover, we analyze the performance of the approach for retrieving top-$k$ candidates instead of only one. 
In the real scenarios, by integrating a linker into a full question answering pipeline, the query generator can process multiple candidates per entity/relation mention (see~\cite{zafar2018formal}), which leads to increase in the performance of the whole QA pipeline (see Section \ref{sec:qa}). 
Hence, we expand the list of generated candidates to consider top-$k$ items and report the performance in terms of MRR in Figure~\ref{fig:mrr_k} for EARL, MDP-Parser and EARL+MDP-Parser (Falcon is omitted as it provides a single candidate per each entity/relation mention). 
The results show that MDP-Parser constantly outperforms EARL for different values of $k$ in both entity linking (a) and relation linking (b) tasks. 
Particularly, we observe a significant enhancement (about 5\%) of our approach in relation linking, which is due to the fact that there are linked items with a very similar label (e.g., \textit{dbr:school} and \textit{dbr:schools}) in the KG. 
This means that by providing more items in the list of candidates, the query generator is at liberty to use the most appropriate candidate.

\subsubsection{Performance of an Overall QA Pipeline}\label{sec:qa}
In order to study the impact of our shallow parsing approach in a complete question answering pipeline, we implement the pipeline shown in Figure~\ref{fig:pipeline}, in which we employ SQG~\cite{zafar2018formal} to serve as the query building component. In this pipeline, the parser annotates the input questions as required by the linker component. The linker component further extracts a list of candidates from the underlying KG for each input from the parser. 
Lastly, the query builder component takes the candidate items from the linker and generates the final candidate formal queries. 

\begin{figure}[t]
\centering
\includegraphics[width=0.30\textwidth]{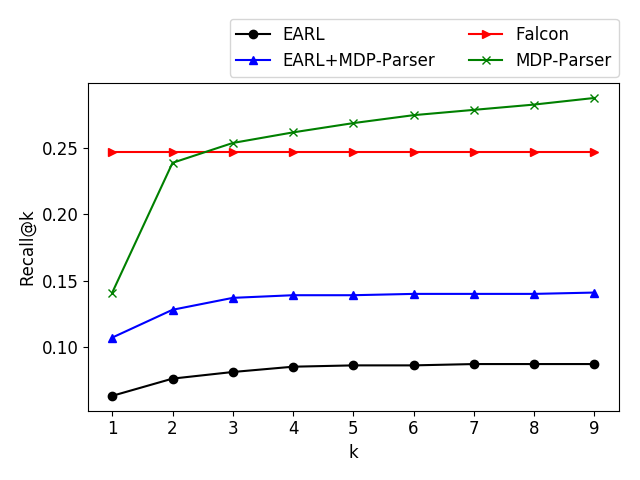}
\caption{The recall of various QA pipelines w.r.t. the value of $k$.}
\label{fig:qa}
\end{figure}

We thus instantiate four realizations of the pipeline: (i) EARL with its internal parsing (ii) Falcon with its internal parsing (iii) combination of MDP-Parser with our ad-hoc linking component explained in Section~\ref{sec:link} (iv) EARL as the linking component with MDP-Parser as parsing component.
Figure~\ref{fig:qa} depicts the performance of each pipeline on the LC-QuAD dataset in terms of recall, which is defined as whether the query builder was able to create the target formal query using top-$k$ candidates by the linker, irrespective of its ranking among all candidate formal queries. We can observe from Figure~\ref{fig:qa} that  the pipeline using MDP-Parser, i.e. (iii), achieves the best score for $k\geq3$. Also, the pipeline that uses MDP-Parser as the shallow parser and EARL as linker (EARL+MDP-Parser) achieves a performance boost of about 5\% in comparison to EARL.

\subsubsection{Case Sensitivity}
In many scenarios, the input questions from the users contain different variety of mistakes including ignoring capitalization.
The lower/upper case usually helps to find out the entities, and many approaches depend on them. 
In this experiment, we show that our approach is not affected as much as others. 
We thus evaluate the robustness of our method, with respect to case sensitivity, compared to two state of the art baselines: EARL and Falcon. 
To do so, we modify all the questions in the LC-QuAD data to lowercase and re-evaluate the methods. 
Table~\ref{tbl:lower_case} shows that the performance of both baselines remarkably falls down on the modified dataset. 
The accuracy of EARL drops by 60\% and 10\%, and Falcon by 29\% and 23\% for entity linking and relation linking tasks, respectively (cmp. Table~\ref{tbl:compare_entity} and Table\ref{tbl:compare_relation}). 
On the other hand, MDP-Parser performs almost the same as before for relation linking (only 1\% decrease).
However, the accuracy of entity linking drops by 15\%. 
This result indicates that while MDP-Parser certainly benefits from capitalization, it does not totally depend on that compared to the baselines. 
\looseness=-1

\begin{table}
\centering
\caption{Accuracies on LC-QuAD in lowercase}
\label{tbl:lower_case}
{\small
\begin{tabular}{@{}lllll@{}}
\toprule
Approach & Entity linking & Relation linking  \\ \midrule
EARL  & 0.006 & 0.164   \\ 
Falcon  & 0.453 & 0.152   \\ 
MDP-Parser & \textbf{0.612} & \textbf{0.443}  \\\bottomrule 
\end{tabular}}

\end{table}

\subsection{Error Analysis}
\begin{figure}[t]
\includegraphics[width=0.30\textwidth]{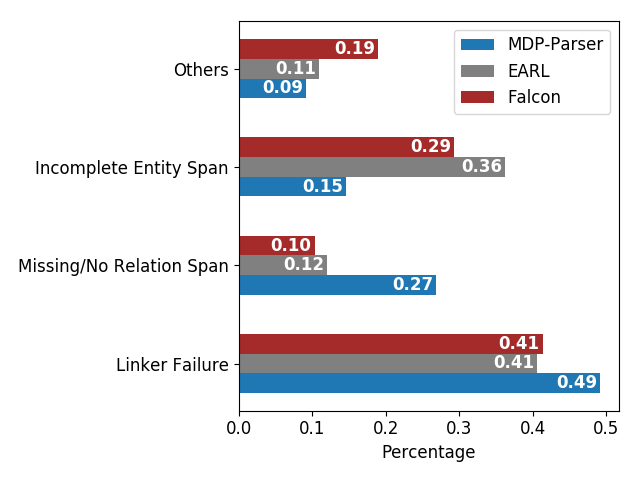}
\caption{Error analysis of the results.}
\label{fig:error_analysis}
\end{figure}

To better understand the success and failure cases of MDP-Parser and the baseline systems, we randomly select 250 out of 1,000 questions from the LC-QuAD test set and perform a thorough manual assessment of failure cases for MDP-Parser, EARL, and Falcon in four categories: Linker Failure, Missing/No Relation Span, Incomplete Entity Span, and others. 
Figure~\ref{fig:error_analysis} depicts the failure percentage of different methods.
The most failure in all three methods occurs because of failure in the linking component, when the linker fails to retrieve the target items, given that the shallow parser provided the correct entity/relation mentions. 
For instance, in the question {\small\texttt{To which places do the flights go by airlines headquartered in the UK}}, the phrase {\small\texttt{UK}} was identified as a potential entity, yet the linker failed to find {\small\texttt{dbr:United\_Kingdom}}. 

The second most common case for MDP-Parser is where it fails to extract at least one relation in the question. As an example, given the question {\small\texttt{Who are the stockholder of the road tunnels operated by the Massachusetts Department of Transportation?}}, with target relation set of {\small\texttt{[dbp:owner, dbp:operator, dbo:RoadTunnel]}}, MDP-Parser correctly identified {\small\texttt{stockholder}} and {\small\texttt{operated}} as candidate relation spans, however, failed to recognize {\small\texttt{road tunnels}}. 
However, the baseline systems tend to offer more relation spans than MDP-Parser, which caused them to falsely suggest the relation mentions that are not related to the target relations. 

The next frequent reason for failure in MDP-Parser is incomplete entity span, which usually occurs for long entity mentions. As an instance, for question {\small\texttt{Who created the Women in the Garden and also the L'Enfant a la tasse?}}, MDP-Parser correctly identified {\small\texttt{L'Enfant a la tasse}} as one entity, however, failed to cast {\small\texttt{Women in the Garden}} as an entity mention, instead it presented {\small\texttt{Woman}} and {\small\texttt{garden}} as two separated entity mentions. Still, MDP-Parser generally handles long entity mentions better than the baseline systems. Lastly, there are less frequent cases such as the instances where two entity/relation mentions reported as one, or where there is no entity mention, etc.

\section{Conclusions}\label{sec:conc} 
In this paper, we designed a shallow parsing method, called MDP-Parser, for question answering systems over knowledge graphs. Considering that the true labels are not available, we presented a sequential approach based on reinforcement learning which is able to model the distantly supervised task by computing a delayed reward signal from distantly computed scores. Furthermore, we combined our parser with an ad-hoc linking component and empirically showed that our method significantly outperforms the baseline systems. In addition, integrating our method into an existing linker remarkably boosted the performance of the linker on the benchmarking datasets.  


\paragraph{Acknowledgements.}\label{sec:Acknowledgments} We acknowledge support by Deutsche Forschungsgemeinschaft (DFG) within the Collaborative Research Center SFB 876 "Providing Information by Resource-Constrained Analysis", project B3 as well as by German Federal Ministry of Education and Research funding for the project SOLIDE (no. 13N14456) and the EU ITN project Cleopatra (grant no. 812997).
\balance
\bibliographystyle{ecai/ecai}
\bibliography{main}

\end{document}